\documentclass[letterpaper, 10 pt, conference]{ieeeconf}  

\IEEEoverridecommandlockouts                              

\usepackage{graphics} 

\usepackage{amsmath}
\usepackage{amssymb}
\usepackage{algorithm}
\usepackage{algpseudocode}
\usepackage{booktabs}
\usepackage{adjustbox}
\usepackage{hyperref}
\usepackage{balance}

\begin{document}

\title{\LARGE
Enhancing Reusability of Learned Skills for Robot Manipulation \\ via Gaze Information and Motion Bottlenecks
}

\author{
Ryo Takizawa$^{1}$, Izumi Karino$^{1}$, Koki Nakagawa$^{1}$, Yoshiyuki Ohmura$^{1}$, and Yasuo Kuniyoshi$^{1}$
\thanks{
This work was supported in part by Grant-in-Aid for Scientific Research(S) JP25H00448 from JSPS.
}%
\thanks{
$^{1}$The authors are with Graduate School of Information Science and Technology, Mechano-Informatics Department,
The University of Tokyo, Japan 
{\tt\footnotesize \{takizawa, karino, k-nakagawa, ohmura, kuniyosh\}@isi.imi.i.u-tokyo.ac.jp}
}%
\thanks{
© 2025 IEEE.  Personal use of this material is permitted.  Permission from IEEE must be obtained for all other uses, in any current or future media, including reprinting/republishing this material for advertising or promotional purposes, creating new collective works, for resale or redistribution to servers or lists, or reuse of any copyrighted component of this work in other works.
}
}



\maketitle

\begin{abstract}
Autonomous agents capable of diverse object manipulations should be able to acquire a wide range of manipulation skills with high reusability. Although advances in deep learning have made it increasingly feasible to replicate the dexterity of human teleoperation in robots, generalizing these acquired skills to previously unseen scenarios remains a significant challenge.
In this study, we propose a novel algorithm, Gaze-based Bottleneck-aware Robot Manipulation (GazeBot), which enables high reusability of learned motions without sacrificing dexterity or reactivity. By leveraging gaze information and motion bottlenecks—both crucial features for object manipulation—GazeBot achieves high success rates compared with state-of-the-art imitation learning methods, particularly when the object positions and end-effector poses differ from those in the provided demonstrations.
Furthermore, the training process of GazeBot is entirely data-driven once a demonstration dataset with gaze data is provided.
Videos and code are available at \url{https://crumbyrobotics.github.io/gazebot}.
\end{abstract}


\section{Introduction} 
Recent
advancements utilizing powerful neural networks such as Transformers have made deep imitation learning increasingly capable of reproducing dexterity to a certain extent \cite{Zhao2023}.
However, significant issues persist regarding their generalization capabilities. 
Although generalization in object manipulation occurs at multiple levels, even the most fundamental aspects, such as changes in object position and the end-effector pose, are known to cause drastic reductions in success rates with variations of just a few centimeters \cite{Cheng2024}.

\begin{figure}[t]
    \centering
    \includegraphics[width=0.95\linewidth]{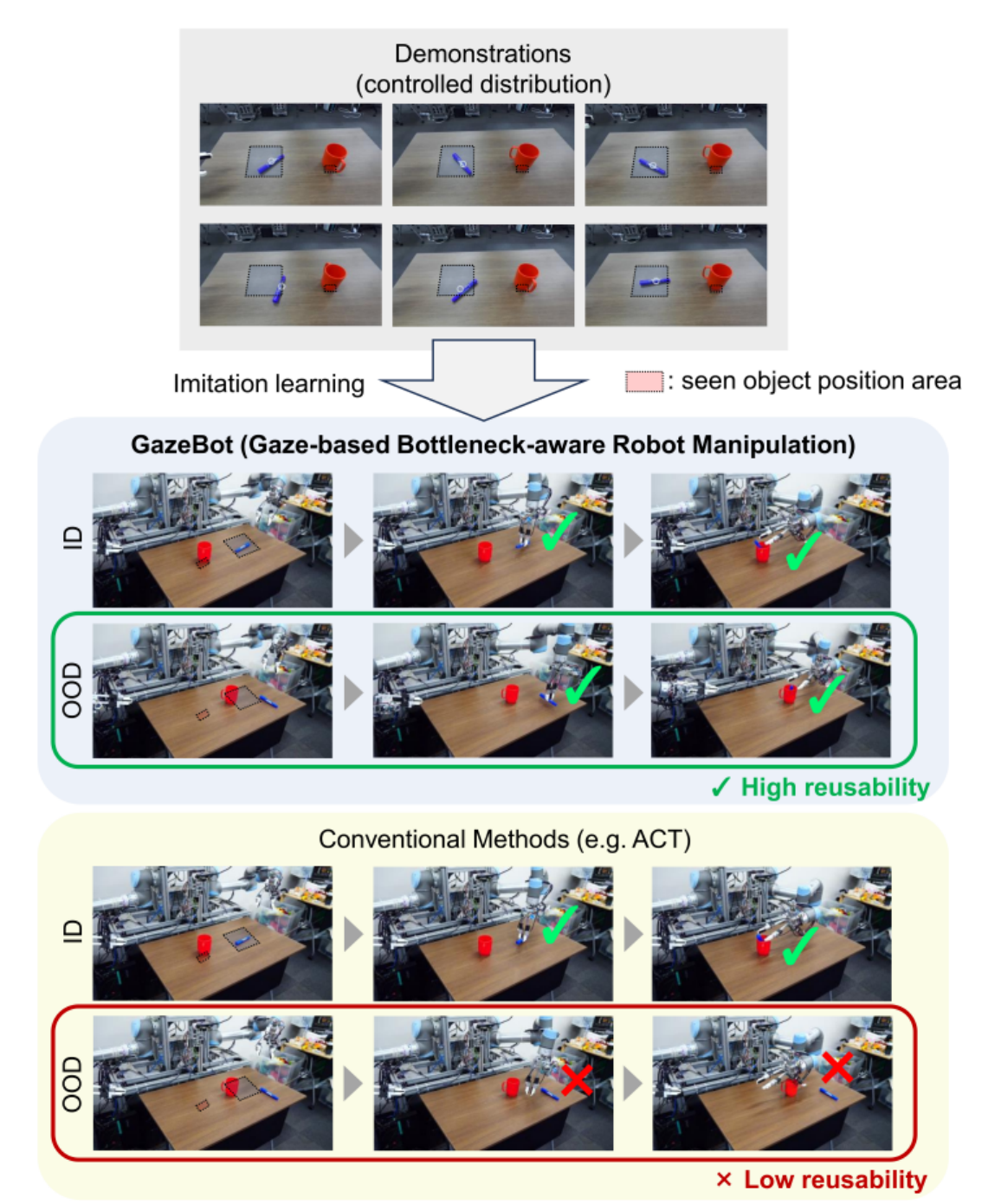}
    \caption{\textbf{GazeBot} achieves high reusability of learned skills for unseen object positions and end-effector poses. Demonstrations collected within restricted ranges of object positions and end-effector poses, and then the success rate is evaluated for in-distribution (\textbf{ID}) cases within these ranges and out-of-distribution (\textbf{OOD}) cases outside them.}
    \label{fig:overview}
\end{figure}

The purpose of this study is to achieve imitation that can accurately perform demonstrated object manipulations under various object positions and initial end-effector poses. 
Owing to the limited generalization capability of conventional imitation learning, exhaustive demonstration collection is currently required to ensure that the robot behaves correctly under various object positions and end-effector poses \cite{Brohan2022, Kim2024b}. 
To address this issue,
we propose Gaze-based Bottleneck-aware Robot Manipulation (\textbf{GazeBot}), an object manipulation imitation method that enables the reuse of acquired skills even with object positions and end-effector poses not included in the provided demonstrations.
As illustrated in Figure \ref{fig:overview}, the object positions and initial end-effector poses used in the demonstrations are restricted to a designated region to evaluate the reusability of the learned skills.
In this setting, GazeBot can accurately perform the task not only under in-distribution (\textbf{ID}) conditions (i.e., within the designated region) but also under out-of-distribution (\textbf{OOD}) conditions (i.e., outside the designated region).


To reuse skills learned within the ID domain in OOD situations, it is necessary to (1) establish an object representation that is robust to changes in object position and (2) develop an action policy architecture capable of accurate control under previously unseen end-effector poses. 
To achieve this, we draw inspiration from human object manipulation, where \textbf{gaze} on the target object not only provides a visual representation independent of the object's absolute position but also exhibits strong gaze–hand coordination during end-effector movements such as reaching \cite{Johansson2001, Hayhoe2003}. 
Concretely, we first represent the entire field of view using a 3D point cloud and then crop a cubic region around the gaze position---referred to as \textbf{gaze-centered point cloud}---from the entire point cloud.
By using this gaze-centered point cloud as input to the action prediction, we realize an object representation that is robust to variations in object position.
Next, based on the action predictivity in the gaze-centered point cloud, we perform a data-driven action segmentation of the overall manipulation into (1) a reaching motion to the vicinity of the gaze position and (2) a gaze-centered action. 
We define the temporal boundary between these motions as a \textbf{bottleneck}, and predict the \textbf{bottleneck pose}, the end-effector pose at this bottleneck, from the 3D gaze position and gaze-centered point cloud. 
This approach allows accurate prediction of the bottleneck pose even for unseen object positions, and this prediction is independent of the current end-effector pose.
Consequently, GazeBot can handle various object positions and end-effector poses by first executing a rough reaching motion to the bottleneck pose, then performing a gaze-centered action using only the gaze-centered point cloud. This approach enables the reuse of the learned motion in unseen conditions. 
Here, reaching the bottleneck is achieved by generating an end-effector trajectory that smoothly connects the current end-effector pose and the bottleneck pose using a first-order Bézier curve, which provides sufficient expressive power for the reaching motion while avoiding unnecessary complexity that could compromise the model's generalization. 
Furthermore, GazeBot updates all actions at every step, and a fully parametric method based on a Transformer is used to directly output subsequent end-effector poses for the gaze-centered actions. 
This approach ensures that dexterity and reactivity are not sacrificed in the pursuit of 
reusability.

\section{Related Work}
\textbf{Gaze-based Object Manipulation.} 
Some works have previously proposed imitation learning methods using gaze data collected from a remote human operator during teleoperated demonstrations for action prediction \cite{Kim2020, Kim2021}. 
These gaze-based methods have exhibited advantages such as enhanced robustness by disregarding task-irrelevant objects \cite{Kim2020} and improved dexterity by focusing on task-relevant regions of visual inputs \cite{Kim2021}. 
However, because these methods rely on image cropping for gaze-centered images, they are susceptible to visual variations caused by changes in object position.
In this study, we employ a gaze-centered point cloud, which provides robustness to positional changes while retaining the conventional benefits of gaze.

\textbf{Data-driven Action Segmentation.} 
Segmenting actions into reaching phase and interaction phase has been proposed to improve dexterity \cite{Kim2024}, increase success rates for long-horizon tasks \cite{Belkhale2023, Sundaresan2024}, and enable high generalization capabilities \cite{Johns2021}. 
Several data-driven segmentation methods have been proposed based on end-effector velocity \cite{Kim2021} or the visibility of the end-effector within a gaze-centered image \cite{Kim2024}. However, these approaches often fail in tasks where high dexterity is not required or when the end-effector is not visible during the interaction phase, such as manipulating with a long stick.
In contrast, our approach segments motions at bottlenecks, which are determined based on action predictivity in the gaze-centered point cloud. 
This action predictivity-based approach offers a more general data-driven segmentation scheme compared with these previous methods.

\textbf{Skill Reusability.}
The reusability of learned skills is required across diverse levels and factors, including adaptation to a variety of object poses and unseen objects in the same category \cite{Goyal2023, Simeonov2021}, to changes in the environment such as varying backgrounds, camera positions, or distractor objects \cite{Zhu2023, Yu2023}, and to entirely novel objects and tasks \cite{Stone2023}. 
However, to the best of our knowledge, no studies have investigated improved
reusability for unseen
object positions and end-effector poses without relying on object- or task-specific assumptions that compromise generality, and without sacrificing dexterity or reactivity.
Our proposed GazeBot is the first deep imitation learning method to demonstrate such reusability under these challenging conditions.
Although several methods exhibit similarities with GazeBot, they have not successfully demonstrated such high reusability, primarily owing to issues in the design of the action policy. 
Hydra \cite{Belkhale2023} and SPHINX \cite{Sundaresan2024}, for instance, segment actions into a reaching motion and a dense action, using sparse action representations similar to a bottleneck pose for the reaching motion. 
However, in contrast to GazeBot, these sparse representations are directly estimated by a neural network from entire images and the end-effector poses, and the dense action prediction also relies on absolute information such as the end-effector poses or entire images. As we will see in Section \ref{sec:experiments}, these design changes hinder the accurate extrapolation of the reaching motions and reduce the reusability of the learned dense actions when faced with unseen object positions or end-effector poses.

\section{Gaze-based Bottleneck-aware Robot Manipulation}
\label{sec:method}

\subsection{Gaze and Bottleneck}
\label{sec:gaze-bottleneck}

\subsubsection{Gaze-based Visual Representation}
Unlike a standard camera, human vision does not uniformly perceive the entire visual field, but instead distinguishes between a high-resolution foveal (central) region and a lower-resolution peripheral region \cite{Paillard1996}. This foveal region can be seen as a 3D attention mechanism that selects a specific portion of space for detailed processing. 
Kim et al. previously proposed an imitation learning method that implements a foveal vision system for robot manipulation \cite{Kim2020}. In that approach, gaze data of a remote human operator are measured during teleoperated demonstrations, and a portion of the image input is cropped around the operator’s gaze position to simulate human-like gaze-based vision. During the inference phase, a gaze prediction model trained on the measured gaze data provides the online gaze control.

\begin{figure}[t]
    \vspace{0.3cm}
    \centering
    \includegraphics[width=0.85\linewidth]{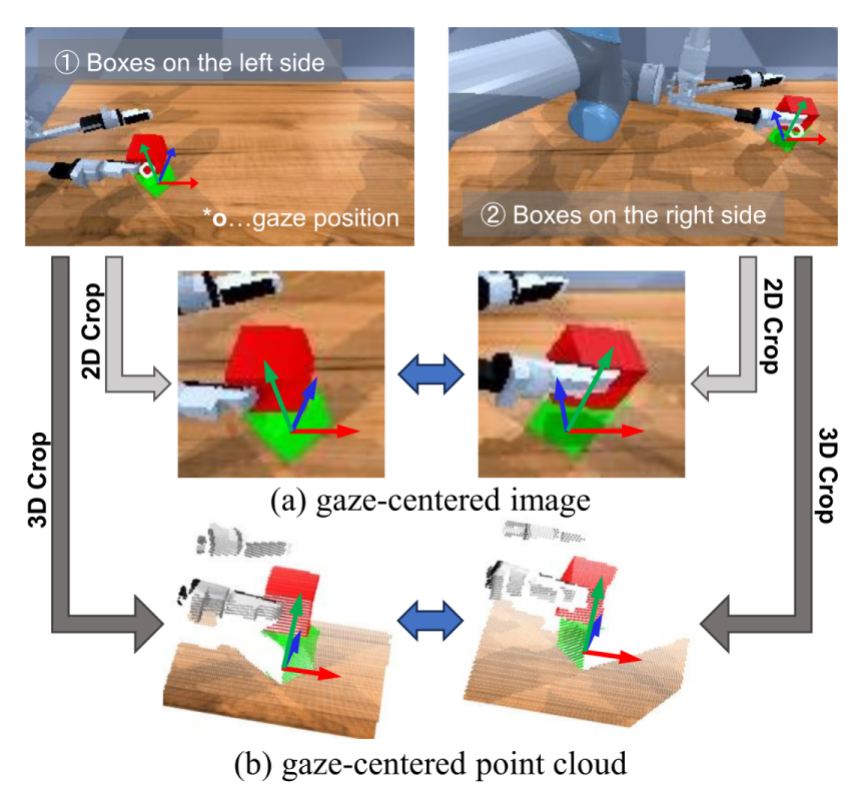}
    \caption{Although both the left and right scenes represent a similar state, their positions on the table differ. In conventional gaze-centered image (a), which lacks 3D-awareness, the scenes appear substantially different, whereas in our proposed \textbf{gaze-centered point cloud} (b), their underlying three-dimensional structure is captured as similar.}
    \label{fig:gaze-vision}
\end{figure}

However, the conventional image-cropping approach lacks 3D-awareness, causing substantial changes in visual representation when the object position is varied (Figure \ref{fig:gaze-vision}).
To address this, we extract a region of the point cloud around the 3D gaze coordinates rather than cropping a 2D image. Here, we use stereo vision to estimate depth and then convert the pixel coordinates of the gaze to 3D coordinates. As shown in Figure \ref{fig:gaze-vision}b, this gaze-centered point cloud reduces the sensitivity to changes in object positions and end-effector poses compared with conventional 2D cropping methods (Figure \ref{fig:gaze-vision}a), thereby enabling neural networks to make predictions independent of object position and end-effector pose.

\subsubsection{Bottleneck-aware Action Segmentation}
\label{sec:bottleneck}

\begin{figure}[t]
    \vspace{0.3cm}
    \centering
    \includegraphics[width=0.9\linewidth]{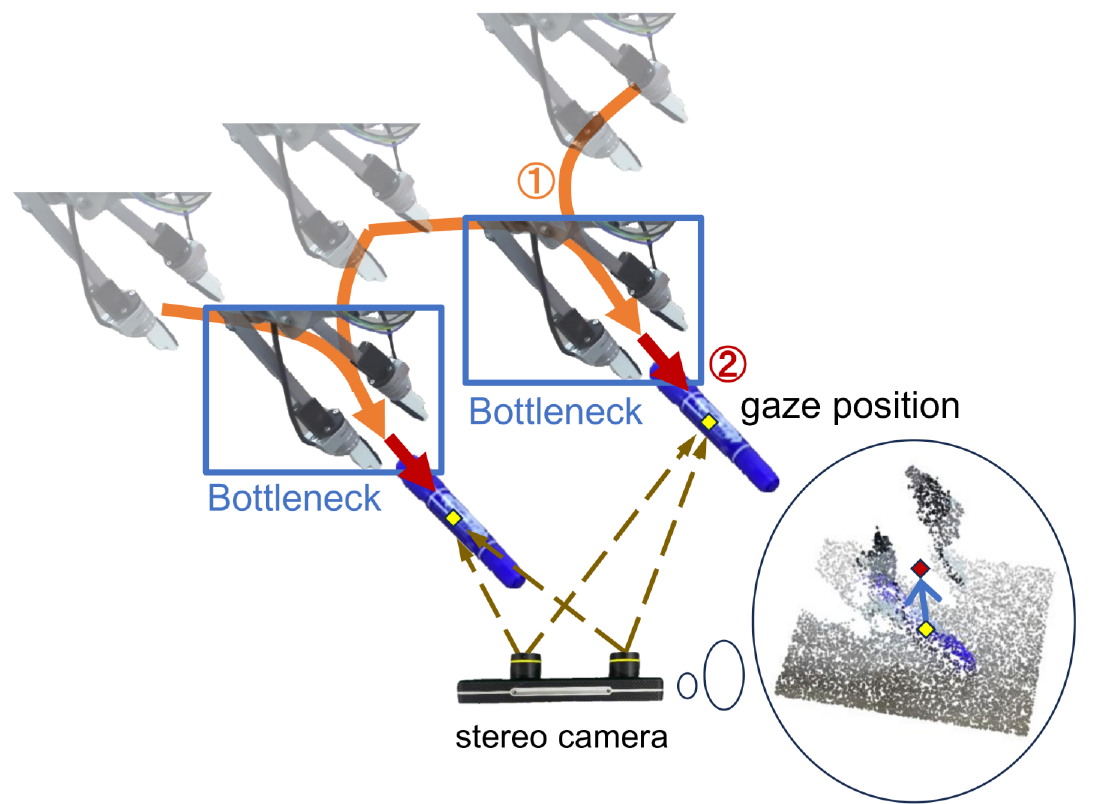}
    \caption{By observing object manipulation in gaze-centered point cloud, an action can be segmented into two phases at the \textbf{bottleneck}: (1) the reaching motion and (2) the gaze-centered action. The motion after the bottleneck is reusable irrespective of the absolute object position and the initial end-effector pose.}
    \label{fig:bottleneck}
\end{figure}

In human object manipulation, gaze position and hand movement are strongly coupled both temporally and spatially \cite{Johansson2001, Paillard1996}.
As illustrated in Figure \ref{fig:bottleneck}, a gaze-centered point cloud makes it possible to segment the movement into two distinct phases: (1) a reaching motion toward the vicinity of the gaze, and (2) a gaze-centered action. This segmentation is possible because gaze-centered vision captures only the object during the reaching phase, whereas it captures both the object and end-effector (or grasped object) during the gaze-centered action phase.
Here, we define the temporal boundary between these two actions as a bottleneck.

While fully parametric methods, in which neural networks directly predict action trajectories, can achieve high dexterity and reactivity, it has been challenging to extrapolate learned behaviors to unseen object positions and end-effector poses.
To address this, we restrict the use of fully parametric action prediction to the gaze-centered actions derived from the aforementioned bottleneck-aware action segmentation. In this context, the robot first reaches the bottleneck pose in a goal-fixed manner using Bézier curve approximation, and then switches to the fully parametric approach for the gaze-centered action.
Here, if (A) the bottleneck estimation is independent of both object position and end-effector pose, and (B) the fully parametric gaze-centered action prediction relies solely on the relative spatial relationships between the object and end-effector (or grasped object), it becomes possible to accurately perform the learned motions even under previously unseen object positions and initial end-effector poses without sacrificing dexterity or reactivity.

As the bottleneck pose is defined in a gaze-centered manner, it can be described as an offset from the 3D gaze position:
\begin{align}
    p_b &= p_{\text{gaze}} + p^{\text{relative}}_b \label{eq:bottleneck1}, \\ 
        &= p_{\text{gaze}} + f(g) \label{eq:bottleneck2}, 
\end{align}
where $p_b, p_{\text{gaze}},$ and $p^{\text{relative}}_b$ denote the bottleneck pose, the 3D gaze position in the end-effector coordinate frame, and the offset of the bottleneck pose from the gaze position, respectively, and $f(\cdot)$ represents a neural network that takes the gaze-centered point cloud $g$ as input.
Here, the bottleneck pose is estimated as an offset from the gaze position rather than being directly predicted based on the gaze-centered point cloud (Eq. \ref{eq:bottleneck2}).
Notably, this offset is independent of the object position. 
As a result, the bottleneck pose can be accurately estimated even when the object position is unseen. Moreover, as Eq. \ref{eq:bottleneck1} does not involve the current end-effector pose, the accuracy of bottleneck estimation is not affected even with unseen end-effector poses.

\subsection{Data-driven Demonstration Segmentation}
\label{sec:segmentations}
Before training an action policy, we first temporally segment the provided demonstrations in a data-driven manner based on gaze and bottlenecks.
Specifically, we decompose the demonstrations into multiple sub-tasks and then further segment each sub-task into reaching motions and gaze-centered actions according to bottlenecks. 
The proposed method for demonstration segmentation is summarized in Algorithm \ref{alg:segmentations}.

\begin{figure}[!t]
\begin{algorithm}[H]
\caption{Data-driven Demonstration Segmentation}
\label{alg:segmentations}
\begin{algorithmic}[1]
  \Statex
  \textbf{Given:} Demonstrations 
  \[
  D = \{(o_t^{(i)}, p_t^{(i)}, g_t^{(i)})\}_{i=1}^{N},\quad t\in[0,T^{(i)}],
  \]
  where $o_t$, $p_t$, $g_t$ denote observation (point cloud), end-effector pose, and gaze position.
  \Statex
  \textbf{Notation:} For the $k$-th sub-task of the $i$-th demonstration: 
  \[
  s^{(i)}_k \text{ (start)},\quad e^{(i)}_k \text{ (end)},\quad b^{(i)}_k \text{ (bottleneck)}
  \]
  
  \State \textbf{(1) Gaze-based Action Segmentation:}
  \For{each demonstration \(i\)}
    \State Partition \([0, T^{(i)}]\) into \([s^{(i)}_0,e^{(i)}_0],\dots,[s^{(i)}_K,e^{(i)}_K]\)
  \EndFor
  
  \State \textbf{(2) Train Action Prediction Model $h_\theta$:}
  \[
  \theta^*=\arg\min_\theta \sum_{(o_t,a_t,g_t) \sim D}\|a_t-h_\theta(\text{crop}(o_t,g_t))\|^2
  \]
  
  \State \textbf{(3) Compute Action Predictivity:}
  \For{each demonstration \(i\) and sub-task \(k\)}
    \State \(\text{scores}_k=\{\|a_t-h_{\theta^*}(\text{crop}(o_t,g_t))\|^2\}_{t=s^{(i)}_k}^{e^{(i)}_k}\)
  \EndFor
  
  \State \textbf{(4) Detect Bottleneck:}
  \For{each sub-task \([s^{(i)}_k,e^{(i)}_k]\)}
    \State Partition \([s^{(i)}_k,e^{(i)}_k]\) into \([s^{(i)}_k, b^{(i)}_k-1]\) and \([b^{(i)}_k, e^{(i)}_k]\)
  \EndFor

  \Statex 
  \textbf{Return:} $D$ and $\{s^{(i)}_k, e^{(i)}_k, b^{(i)}_k\}_{i,k}$
\end{algorithmic}
\end{algorithm}
\end{figure}

\subsubsection{Gaze-Based Sub-task Segmentation}
\label{sec:sub-task}
A typical object manipulation task comprises multiple bottlenecks. For example, in a standard pick-and-place task, there is usually one bottleneck for the “pick” phase and another for the “place” phase. In such cases, before segmenting the motions using bottlenecks, it is necessary to decompose a sequence of object manipulation behaviors into multiple smaller sub-tasks (e.g., pick/place in this example) so that each sub-task includes exactly one bottleneck.

As a robust and simple way to achieve such segmentation in a data-driven manner, previous work proposed a gaze-based task decomposition method for object manipulation \cite{Takizawa2024}. 
This approach exploits the pattern of gaze during object manipulation, where the human teleoperator fixates on specific task-relevant gaze landmarks. By simply detecting transitions in this gaze behavior, the task can be segmented so that each fixation period corresponds to a distinct sub-task. In this work, we first apply this method to decompose expert demonstrations into multiple sub-tasks.

\subsubsection{Bottleneck Determination}
\label{sec:bottleneck-determination}
Once the demonstrations are segmented into sub-tasks, we need to determine the time step at which each sub-task reaches its bottleneck pose. 
To do so, we propose a bottleneck determination approach based on action predictivity from a gaze-centered point cloud.
In each sub-task, predicting actions solely from a gaze-centered point cloud exhibits low accuracy when the end-effector or grasped object is not visible within the gaze-centered point cloud.
Therefore, the transition from this low-predictivity phase to a high-predictivity phase can serve as the segmentation point, which we define as the bottleneck. 

In this approach, we use a policy model $a_t = h(g_t)$ that predicts the action solely from the gaze-centered point cloud $g_t$. This policy model is trained via behavior cloning \cite{Zhang2018}, that is, by minimizing $\|h(g_t) - a_t^*\|$ to replicate the actions in the demonstration, where $a_t^*$ denotes the expert action recorded in the demonstration. 
Once this policy $h$ has been trained, we compute the action-prediction loss $\|h(g_t) - a_t^*\|$ at every time step of each demonstration. 
Finally, we split each sub-task based on the median of this action prediction loss, enabling entirely data-driven segmentation for every sub-task.


\subsection{Policy Design}
\label{sec:model}

\begin{figure*}[t]
    \vspace{0.3cm}
    \begin{center}
        \includegraphics[width=\linewidth]{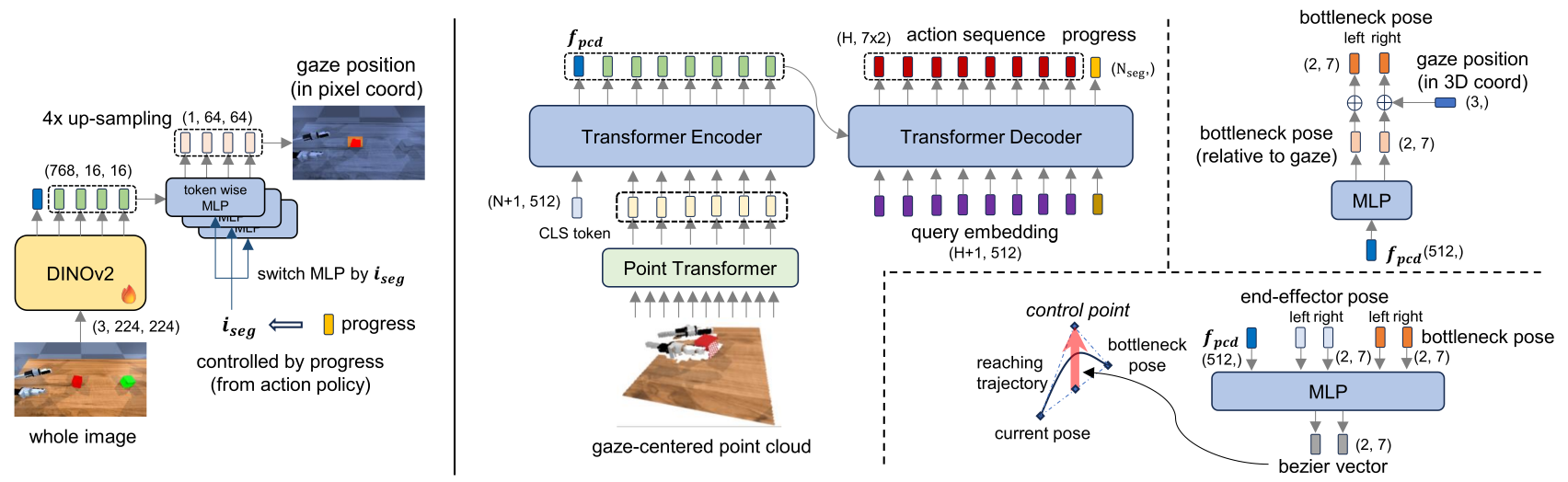}
        \caption{\textbf{GazeBot} architecture. (left) The gaze prediction model is trained to estimate the gaze position across the entire image as a classification problem. (right) The action policy model achieves robust reaching motion by estimating the bottleneck pose and the shape of the trajectory up to the bottleneck, and uses a Transformer to predict the gaze-centered action in a full-parametric manner. Both actions and gaze transitions are predicted by the gaze-centered point cloud to improve the reusability.}
        \label{fig:model-architecture}
    \end{center}
\end{figure*}

At every time step, GazeBot first predicts the gaze position from the entire image using the gaze prediction model. It then uses the gaze-centered point cloud at the predicted gaze position, along with the current end-effector pose, as inputs to the action policy model, which reactively predicts the subsequent end-effector poses.

\subsubsection{Gaze Prediction Model}  
\label{sec:gaze-model}
For the gaze prediction model, a human teleoperator’s gaze is measured simultaneously during demonstration collection, and this gaze data is then used as supervision \cite{Kim2020}. 
The gaze positions estimated by the model are used during both training and inference.
As shown in the left side of Figure \ref{fig:model-architecture}, the image embeddings are extracted via DINOv2 \cite{Oquab2023}, and each 768-dimensional embedding is mapped to a single probability value with a four-layer multilayer perceptron (MLP).

To intentionally transition the gaze to the next gaze landmark upon completing each sub-task, we prepare one such MLP for each sub-task and switch these MLPs based on the sub-task index $i_{\mathrm{seg}}$.  
At inference time, the sub-task index $i_{\mathrm{seg}}$ is initialized to zero at the beginning and incremented based on the progress $c_t$ output by the action policy model, indicating the completion of each sub-task. 
By managing gaze transitions using $c_t$ estimated from the gaze-centered point cloud, which solely captures the relative spatial relationships between the object and end-effector (or grasped object), the robot can perform gaze transitions at the correct timing even under unseen conditions.

\subsubsection{Action Policy Model (i)}  
\label{sec:policy-model1}
The pixel-space gaze position predicted by the gaze prediction model is transformed into 3D coordinates using stereo-based depth estimation. 
From the stereo-derived point cloud, we then crop a cubic region (20 cm on each side in this study) centered on the 3D gaze coordinates, which is referred to as the gaze-centered point cloud. 
This gaze-centered point cloud is embedded as a sequence of tokens using a PointTransformer \cite{Zhao2020} and then passed to a Transformer encoder (Figure \ref{fig:model-architecture} right). 
The Transformer encoder includes a CLS token (\textit{classification token}, similar to ViT \cite{Dosovitskiy2020}) that aggregates information from all point-cloud tokens into a feature vector $f_{\mathrm{pcd}} \in \mathbb{R}^{512}$. 
All output tokens, including the CLS token, are then fed into a Transformer decoder, which generates both the action sequence $a_{t:t+H} \in \mathbb{R}^{H\times14}$ for the left and right end-effector over $H$ future time steps and the progress $c_t \in \mathbb{R}^{N_{seg}}$ for controlling gaze transitions in each sub-task. Here, each end-effector has 7 degrees of freedom including gripper angle, and $N_{seg}$ denotes the number of sub-tasks. 
As the input vision is gaze-centered, the output actions are represented as relative end-effector poses, with the current end-effector pose serving as the origin of the coordinate frame. 
By computing these actions in relative coordinates, the learned gaze-centered actions remain reusable even when the absolute position of the object changes.

\subsubsection{Action Policy Model (ii)}  
\label{sec:policy-model2}
The feature vector $f_{\mathrm{pcd}}$ is used to predict the bottleneck poses for each end-effector. 
As described in Section \ref{sec:bottleneck}, the bottleneck pose is estimated by first predicting its offset relative to the 3D gaze position, then adding this offset to the 3D coordinates of the gaze. The module depicted in the top-right of Figure \ref{fig:model-architecture} implements this procedure and enables significantly more accurate extrapolation of the bottleneck pose under unseen object positions.

After predicting the bottleneck pose, the corresponding reaching trajectory is generated. 
Because this trajectory is non-contact and relatively coarse, we model it with a first-order Bézier curve, capturing its overall shape without introducing unnecessary complexity that could hinder generalization.
As illustrated in the lower-right portion of Figure \ref{fig:model-architecture}, the model predicts a 7-dimensional \textit{bezier vector} that determines the shape of the first-order Bézier curve for each end-effector.
A first-order Bézier curve is defined by its two endpoints and a single \textit{control point}. 
Rather than predicting the control point directly, we first estimate the displacement---bezier vector---from the mean pose of the start (current end-effector pose) and end (bottleneck pose) poses, and then add this bezier vector to the mean pose to obtain the control point (Figure \ref{fig:model-architecture} bottom-right).
For the training, we approximate each reaching motion in the demonstrations by fitting them with first-order Bézier curves, and then use the resulting bezier vectors from these curves for supervision.

\section{Experiments}
\label{sec:experiments}
\subsection{Robot System}
We used a dual-arm robot system designed for imitation learning via human teleoperation. 
In this system, a human operator remotely controls the robot while observing its surrounding environment through a head-mounted display (HMD). During this operation, the operator’s gaze data were recorded in sync with the video feed displayed on the HMD. The system is compatible with both a physical dual-arm robot, consisting of two UR5 (Universal Robots Inc.) arms, and its simulated counterpart, allowing the collection of teleoperated demonstration data in both real and virtual environments. 
The recorded gaze data were output as pixel coordinates corresponding to the images displayed on the HMD. The same single stereo camera (ZED Mini, Stereolabs Inc.) was used for both the teleoperation and inference phases, and we directly utilized the depth images generated by the deep learning–based depth estimation algorithm provided by ZED SDK (Stereolabs Inc.). The time-series demonstration data used for training were recorded at 10 Hz.

\begin{figure*}
    \vspace{0.3cm}
    \begin{center}
        \includegraphics[width=\linewidth]{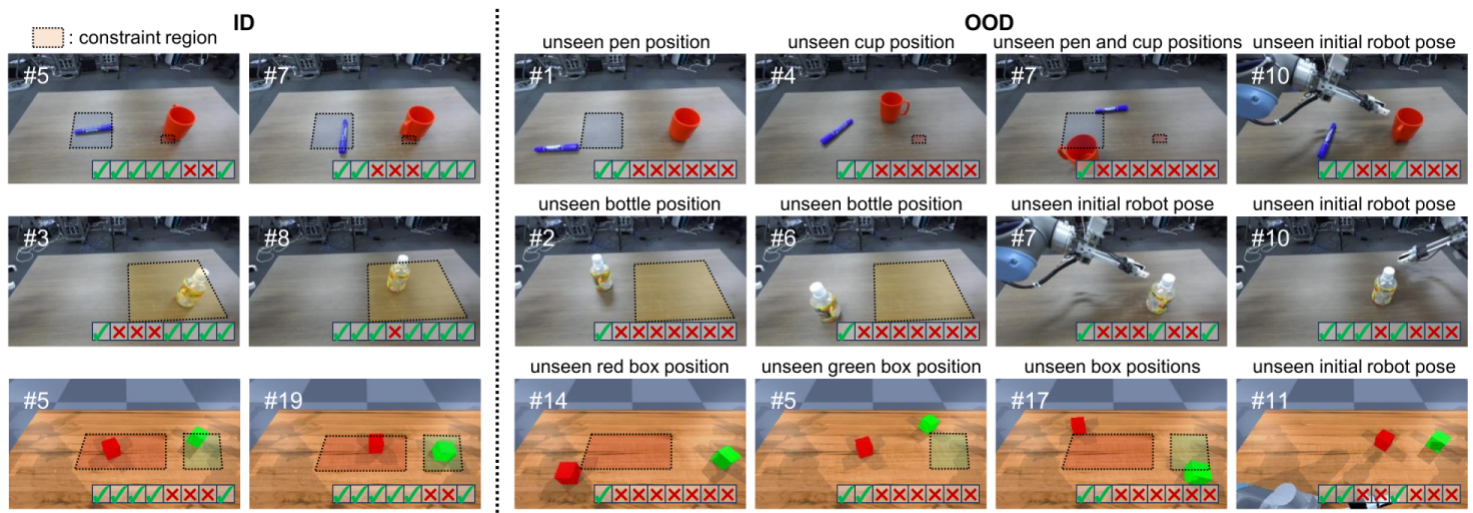}
        \caption{Examples of \textbf{ID} and \textbf{OOD} trials in the PenInCup, OpenCap, and PileBox, where object positions and the initial end-effector poses are controlled. The images show the initial states of each trial. The checkboxes correspond to the method order in Table \ref{tab:eval-generalization} and indicate whether each method succeeded in the task from that initial state.}
        \label{fig:eval-generalization}
    \end{center}
\end{figure*}

\subsection{Task Setup}
\label{sec:data}
We conducted imitation learning experiments on five tasks (four real and one simulated) to evaluate the skill reusability and applicability of GazeBot.

For the first three tasks, to quantitatively assess the 
reusability, we collected demonstrations under controlled conditions by restricting the object positions and initial end-effector poses within predefined regions, thus clearly defining ID and OOD situations:

\begin{itemize}
    \item \textbf{PenInCup} (real, 109 demos): On a bare tabletop, the robot picks up a blue marker pen with its left arm and places it into a red cup. Gaze-based action segmentation splits this into two sub-tasks. The pen is placed with its uncapped end within a 15 cm square on the table's left, and the cup base overlaps a 5 cm area on the right. The robot starts with end-effectors above the table edges, oriented inward. Object and robot poses are manually randomized within defined regions; other factors (e.g., table location, background) are kept consistent. Placement boundaries are subtly marked to avoid visual bias.
    
    \item \textbf{OpenCap} (real, 110 demos): On a bare tablecloth, the robot holds an upright plastic bottle with its right arm and unscrews the cap with its left. The task is divided into two sub-tasks. The bottle is always on the right side of the table. Other settings match PenInCup.
    
    \item \textbf{PileBox} (sim, 100 demos): On a bare tabletop, the robot stacks a red box (picked up with its left arm) onto a green box. The task is divided into two sub-tasks. The red box is placed within a 10×20 cm area on the left, and the green box within a 10 cm square on the right. Other settings match PenInCup.
\end{itemize}

For the remaining two tasks, we chose more complex, long-horizon tasks involving deformable objects to evaluate the model’s applicability (task configurations were approximately randomized):
\begin{itemize}
    \item \textbf{WipeTray} (real, 100 demos): On a green tablecloth, the robot folds a towel, regrasp it, and wipes a tray with it, then places the towel back on the table. The task is divided into five sub-tasks.
    \item \textbf{BaggingGoods} (real, 91 demos): On a green tablecloth, the robot grasps the handle of a bag with its left arm and opens the bag. It then uses its right arm to place two of the three objects on the table into the bag in a specified order. The task is divided into five sub-tasks.
\end{itemize}

\subsection{Evaluation of Reusability and Ablation Studies}

\begin{table*}
    \vspace{0.3cm}
    \centering
    \caption{Comparison of success rates for ID and OOD trials in PenInCup, OpenCap and PileBox. 
    }
    \resizebox{\textwidth}{!}{
    \begin{tabular}{lcccccccccccc}
    \toprule
        & \multicolumn{4}{c}{\textbf{PenInCup} (real)} 
        & \multicolumn{4}{c}{\textbf{OpenCap} (real)} 
        & \multicolumn{4}{c}{\textbf{PileBox} (sim)} \\
    \cmidrule(lr){2-5}\cmidrule(lr){6-9}\cmidrule(lr){10-13}
        & \multicolumn{2}{c}{ID} & \multicolumn{2}{c}{OOD} 
        & \multicolumn{2}{c}{ID} & \multicolumn{2}{c}{OOD} 
        & \multicolumn{2}{c}{ID} & \multicolumn{2}{c}{OOD} \\
    \cmidrule(lr){2-3}\cmidrule(lr){4-5}
    \cmidrule(lr){6-7}\cmidrule(lr){8-9}
    \cmidrule(lr){10-11}\cmidrule(lr){12-13}
        \textbf{Method} 
        & Pick & Put & Pick & Put 
        & Hold & Open 
        & Hold & Open 
        & Lifted & Pile 
        & Lifted & Pile \\
    \midrule
        GazeBot (Ours) 
            & \textbf{11/12} & \textbf{9/12} & \textbf{10/12} & \textbf{10/12} 
            & \textbf{12/12} & \textbf{10/12} 
            & \textbf{12/12} & \textbf{9/12} 
            & \textbf{20/20} & \textbf{20/20} 
            & 16/20 & \textbf{15/20} \\
        - w/o point cloud (Ablation1) 
            & 9/12 & \textbf{9/12} & 5/12 & 5/12 
            & \textbf{12/12} & 8/12 
            & 7/12 & 5/12 
            & 19/20 & 13/20 
            & 14/20 & 8/20 \\
        - w/ state input partially (Ablation2)
            & 6/12 & 4/12 & 3/12 & 1/12 
            & \textbf{12/12} & 6/12
            & 7/12 & 5/12 
            & \textbf{20/20} & 18/20 
            & \textbf{17/20} & 10/20 \\
        - w/ state input (Ablation3)
            & 9/12 & 8/12 & 2/12 & 0/12 
            & \textbf{12/12} & 6/12 
            & 6/12 & 4/12 
            & \textbf{20/20} & 16/20 
            & 13/20 & 2/20 \\
        - w/o relative bottleneck (Ablation4)
            & 7/12 & 6/12 & 5/12 & 2/12 
            & \textbf{12/12} & 8/12 
            & 8/12 & 5/12 
            & \textbf{20/20} & 19/20 
            & 13/20 & 7/20 \\
        DAA 
            & 8/12 & 6/12 & 2/12 & 0/12 
            & 9/12 & 8/12 
            & 1/12 & 1/12 
            & \textbf{20/20} & 12/20
            & 3/20 & 0/20 \\
        - w/o local action 
            & 8/12 & 6/12 & 2/12 & 0/12 
            & 8/12 & 6/12 
            & 1/12 & 1/12 
            & \textbf{20/20} & 8/20 
            & 4/20 & 0/20 \\
        ACT 
            & 9/12 & 8/12 & 1/12 & 0/12 
            & \textbf{12/12} & 9/12 
            & 3/12 & 3/12 
            & \textbf{20/20} & 16/20 
            & 8/20 & 3/20 \\
    \bottomrule
    \end{tabular}
    }
    \label{tab:eval-generalization}
\end{table*}

We first trained GazeBot and its ablation models, including conventional models (ACT \cite{Zhao2023}, DAA \cite{Kim2024b}), using the demonstrations collected as previously described. 
ACT is known for its strong performance and can be considered equivalent to GazeBot with all proposed modules removed. DAA extends ACT by incorporating gaze-based image cropping and action segmentation into global and local actions. DAA w/o local actions can be interpreted as applying only gaze-based vision to ACT.
We then measured their success rates for scenarios within the training distribution (ID) and those outside it (OOD), as shown in Figure \ref{fig:eval-generalization}. 
In the experiment, we standardized the initial conditions across all trials to ensure that all models performed the task under as similar conditions as possible.
In the OOD trials, we first evaluated the cases where each object was individually placed in unseen positions, and then examined the cases where all objects were simultaneously placed in unseen positions (Figure \ref{fig:eval-generalization}). Finally, we conducted trials involving unseen initial end-effector poses. 

The overall success rates in ID and OOD trials for the four tasks are presented in Table \ref{tab:eval-generalization}.
Here, the reusability of acquired skills is evaluated by how well the model achieves a success rate as high as possible on OOD while maintaining a high success rate on ID.
GazeBot incorporates several design choices and components, whose contributions can be assessed by comparing its performance with that of the ablation models.

\subsubsection{3D Awareness of Gaze-based Vision (Ablation1)}
One core feature of GazeBot is the ability to produce visual representations robust to changes in object location, achieved by the gaze-centered point cloud. In Ablation1, we replaced the gaze-centered point cloud with the conventional image-cropping approach. The cropped left and right images were tokenized via ResNet18 \cite{He2015}, following ACT, and then fed into the Transformer encoder. 
Although Ablation1 maintained a relatively high success rate for ID trials, as presented in Table \ref{tab:eval-generalization}, it suffered from lower accuracy in OOD owing to the absence of 3D awareness, which led to degraded performance when the objects appeared differently under unseen object positions.

\subsubsection{Inputs to the Policy (Ablation2, Ablation3)}
In GazeBot, the action policy model is designed so that the current end-effector pose is used as input only during the estimation of the bezier vector. 
In other words, bottleneck estimation relies solely on the 3D gaze position and the gaze-centered point cloud, and the prediction of the gaze-centered actions after the bottlenecks also relies only on the gaze-centered point cloud.

In Ablation3, we added tokens representing the current left and right end-effector poses and the 3D gaze position (3 × 512-dimensions) to the sequence of 3D point cloud tokens fed to the Transformer encoder. This addition prevented the model from accurately estimating the bottleneck in unseen initial poses. Even when the robot successfully reached the correct bottleneck for unseen object positions, the subsequent gaze-centered actions degraded because the bottleneck pose itself is an untrained input for the policy network (Table \ref{tab:eval-generalization}).

In Ablation2, we similarly added tokens but then applied an attention mask so that the CLS token $f_{pcd}$ in the Transformer encoder did not attend to these additional tokens. As a result, $f_{pcd}$ aggregated only information from the 3D point cloud, enabling accurate bottleneck reaching for unseen end-effector poses. Nonetheless, as mentioned in Ablation3, the accuracy of the gaze-centered actions still decreased even after the bottleneck was reached correctly (Table \ref{tab:eval-generalization}), often causing the end-effector to be “pulled” back toward the ID region.

\subsubsection{Bottleneck Estimation Method (Ablation4)}
GazeBot estimates the bottleneck pose by first predicting an offset from the 3D gaze position based on the gaze-centered point cloud and then adding that offset to the 3D gaze position. Another implementation could be considered, where these inputs are fed into a neural network that directly outputs the bottleneck pose. However, under unseen object positions, Ablation4 struggled to extrapolate the bottleneck and even failed to reach the object correctly. By contrast, our method successfully extrapolated the bottleneck pose in almost all OOD cases (Table \ref{tab:eval-generalization}).

\subsubsection{Design of the Bottleneck Reaching Module (DAA)}
In GazeBot, we achieve bottleneck reaching by first predicting a bottleneck pose and then generating a first-order Bézier curve as a trajectory leading to the bottleneck.
One could adopt a fully parametric approach that directly outputs an action sequence leading to the bottleneck pose, as in DAA.
In our tasks, however, DAA exhibited a significant drop in success rates from ID to OOD (Table \ref{tab:eval-generalization}). In most OOD trials, the model even failed to perform the reaching motion itself, reflecting the issues seen in Ablation4.

\subsubsection{Gaze-Based Vision (ACT, DAA w/o local action)}
As can be observed from Table \ref{tab:eval-generalization}, both ACT and DAA exhibited a sharp decline in success rates when moving from ID to OOD.
In OOD trials, whether local action was present or not had little impact on the success rate of DAA because it already failed during the global (reaching) phase.
Moreover, ACT slightly outperformed DAA overall. 
In DAA, because only a small, gaze-centered region is input to the model and 3D awareness is nearly absent, the input varies significantly with the object position, undermining data efficiency. Consequently, with approximately 100 demonstrations, as in this experiment, it was not possible to achieve the improvements in dexterity previously reported \cite{Kim2021}.
Meanwhile, GazeBot, which also adopted gaze-based vision, outperformed ACT (Table \ref{tab:eval-generalization}). This demonstrates that our approach additionally contributed to the improved data efficiency of gaze-based models.

\begin{table}
    \centering
    \caption{Comparison of success rates in WipeTray and BaggingGoods.}
    \begin{tabular}{lccccc}
    \toprule
        & \multicolumn{2}{c}{\textbf{WipeTray} (real)} 
        & \multicolumn{3}{c}{\textbf{BaggingGoods} (real)}  \\
    \cmidrule(lr){2-3}\cmidrule(lr){4-6}
        Method 
        & Folded & Wipe  
        & Open & Insert1 & Insert2 \\
    \midrule
        GazeBot (Ours) 
            & 17/20 & 16/20 & \textbf{17/20} & \textbf{12/20} &  \textbf{11/20}
            \\
        ACT 
            & \textbf{18/20} & \textbf{18/20} & 15/20 & 7/20 & 7/20
            \\
    \bottomrule
    \end{tabular}
    \label{tab:eval-applicability}
\end{table}

\section{Evaluation of Applicability}
We evaluated skill reusability using simple rigid-body tasks to establish a clear ID/OOD split. 
However, the aim of our study is to enhance reusability without limiting the task scope to basic pick-and-place operations involving only rigid objects. 
Therefore, in this section, we demonstrate that GazeBot can also successfully learn more complex and realistic tasks, WipeTray and BaggingGoods, both of which involve long-horizon and deformable-object manipulation.
Tests were conducted under demonstration-like configurations without defining ID/OOD.
As a baseline, we used ACT, one of the most capable model-free approaches.
In each trial, we ensured that all models operated under the same configuration as much as possible.

The results are summarized in Table \ref{tab:eval-applicability}. 
GazeBot successfully learned all tasks, similar to ACT. 
Moreover, it clearly outperformed ACT in BaggingGoods, likely because this task involves many objects and large variations in their configurations. 
In such cases, GazeBot’s gaze- and bottleneck-based canonicalization helps improve learning accuracy. 
In contrast, ACT performed better in WipeTray, where the object is large and its position varies little. 
In these scenarios, ACT can more easily reproduce the demonstrated behavior by directly incorporating proprioception into the action policy.

\section{Limitations and Future Directions}
In this study, we proposed GazeBot, an imitation learning method that significantly improves skill reusability without sacrificing dexterity or reactivity, particularly for unseen object positions and end-effector poses.
However, several challenges remain.

\textbf{Precise and Flexible Gaze Control.} While GazeBot demonstrated the importance of three-dimensional gaze control in object manipulation, robotic gaze prediction is still in its infancy. 
Robots not only require task-specific gaze data from human teleoperators for supervision but also lack the flexibility to adjust their gaze positions when objects are partially or fully obscured. Moreover, they struggle to align their gaze \textit{exactly} on tiny targets, such as a needle in midair.

\textbf{More Flexible and Adaptive Segmentations.} Although we achieved highly reusable imitation by employing gaze and bottlenecks to spatially and temporally segment vision and action, our current approach uses rigid segmentation along explicit boundaries. One limitation concerns the gaze-centered point cloud. In our method, gaze-based vision only captures information corresponding to foveal vision, whereas human vision simultaneously benefits from peripheral vision, which can roughly capture a wider range of information. Another limitation involves bottleneck determination. The proposed approach assumes that each sub-task has a single bottleneck, yet some tasks involve sub-tasks with multiple bottlenecks. Future work could explore more flexible segmentation methods to enable more adaptive learning.

\textbf{Integrating Advanced Trajectory Planning.} For simplicity, the reaching trajectories in this work are generated by connecting the current end-effector pose and the bottleneck pose with a first-order Bézier curve. To further enhance applicability, we could replace these simple connections with trajectories provided by advanced planning methods that support collision avoidance and bimanual coordination. Such improvements would enable more robust reusability of learned motions.


\begin{thebibliography}{10}
\providecommand{\url}[1]{#1}
\csname url@rmstyle\endcsname
\providecommand{\newblock}{\relax}
\providecommand{\bibinfo}[2]{#2}
\providecommand\BIBentrySTDinterwordspacing{\spaceskip=0pt\relax}
\providecommand\BIBentryALTinterwordstretchfactor{4}
\providecommand\BIBentryALTinterwordspacing{\spaceskip=\fontdimen2\font plus
\BIBentryALTinterwordstretchfactor\fontdimen3\font minus \fontdimen4\font\relax}
\providecommand\BIBforeignlanguage[2]{{%
\expandafter\ifx\csname l@#1\endcsname\relax
\typeout{** WARNING: IEEEtran.bst: No hyphenation pattern has been}%
\typeout{** loaded for the language `#1'. Using the pattern for}%
\typeout{** the default language instead.}%
\else
\language=\csname l@#1\endcsname
\fi
#2}}

\bibitem{Zhao2023}
T.~Z. Zhao, V.~Kumar, S.~Levine, and C.~Finn, ``Learning fine-grained bimanual manipulation with low-cost hardware,'' in \emph{Proc. Robot.: Sci. Syst}, 2023.

\bibitem{Cheng2024}
X.~Cheng, J.~Li, S.~Yang, G.~Yang, and X.~Wang, ``Open-television: Teleoperation with immersive active visual feedback,'' in \emph{Proc. Conf. Robot Learn}, 2024.

\bibitem{Brohan2022}
A.~Brohan, \emph{et~al.}, ``Rt-1: Robotics transformer for real-world control at scale,'' in \emph{Proc. Robot.: Sci. Syst}, 2023.

\bibitem{Kim2024b}
H.~Kim, Y.~Ohmura, and Y.~Kuniyoshi, ``Multi-task real-robot data with gaze attention for dual-arm fine manipulation,'' in \emph{Proc. {IEEE}/{RSJ} Int. Conf. Intell. Robots Syst.}, 2024.

\bibitem{Johansson2001}
R.~Johansson, G.~Westling, A.~Bäckström, and J.~Flanagan, ``Eye–hand coordination in object manipulation,'' \emph{J. Neurosci.}, vol.~21, pp. 6917--32, 10 2001.

\bibitem{Hayhoe2003}
M.~M. Hayhoe, A.~Shrivastava, R.~Mruczek, and J.~B. Pelz, ``Visual memory and motor planning in a natural task,'' \emph{J. Vis.}, vol.~3, pp. 49--63, 2003.

\bibitem{Kim2020}
H.~Kim, Y.~Ohmura, and Y.~Kuniyoshi, ``Using human gaze to improve robustness against irrelevant objects in robot manipulation tasks,'' \emph{{IEEE} Robot. Automat. Lett.}, vol.~5, pp. 4415--4422, 7 2020.

\bibitem{Kim2021}
------, ``Gaze-based dual resolution deep imitation learning for high-precision dexterous robot manipulation,'' \emph{{IEEE} Robot. Automat. Lett.}, vol.~6, pp. 1630--1637, 4 2021.

\bibitem{Kim2024}
------, ``Goal-conditioned dual-action imitation learning for dexterous dual-arm robot manipulation,'' \emph{{IEEE} Trans. Robot.}, vol.~40, pp. 2287--2305, 2024.

\bibitem{Belkhale2023}
S.~Belkhale, Y.~Cui, and D.~Sadigh, ``Hydra: Hybrid robot actions for imitation learning,'' in \emph{Proc. Conf. Robot Learn}, 2023.

\bibitem{Sundaresan2024}
P.~Sundaresan, H.~Hu, Q.~Vuong, J.~Bohg, and D.~Sadigh, ``What's the move? hybrid imitation learning via salient points,'' in \emph{Proc. Int. Conf. Learn. Represent.}, 2024.

\bibitem{Johns2021}
E.~Johns, ``Coarse-to-fine imitation learning: Robot manipulation from a single demonstration,'' in \emph{Proc. {IEEE} Int. Conf. Robot. Automat.}, 2021.

\bibitem{Goyal2023}
A.~Goyal, J.~Xu, Y.~Guo, V.~Blukis, Y.-W. Chao, and D.~Fox, ``Rvt: Robotic view transformer for 3d object manipulation,'' \emph{CoRL}, 2023.

\bibitem{Simeonov2021}
A.~Simeonov, Y.~Du, A.~Tagliasacchi, J.~B. Tenenbaum, A.~Rodriguez, P.~Agrawal, and V.~Sitzmann, ``Neural descriptor fields: Se(3)-equivariant object representations for manipulation,'' in \emph{Proc. {IEEE} Int. Conf. Robot. Automat.}, 2021.

\bibitem{Zhu2023}
Y.~Zhu, Z.~Jiang, P.~Stone, and Y.~Zhu, ``Learning generalizable manipulation policies with object-centric 3d representations,'' in \emph{Proc. Conf. Robot Learn}, 2023.

\bibitem{Yu2023}
T.~Yu, \emph{et~al.}, ``Scaling robot learning with semantically imagined experience,'' in \emph{Proc. Robot.: Sci. Syst}, 2023.

\bibitem{Stone2023}
A.~Stone, \emph{et~al.}, ``Open-world object manipulation using pre-trained vision-language models,'' in \emph{Proc. Conf. Robot Learn}, 2023.

\bibitem{Paillard1996}
J.~Paillard, ``Fast and slow feedback loops for the visual correction of spatial errors in a pointing task: a reappraisal.'' \emph{Can. J. Physiol. Pharmacol.}, vol. 74(4), pp. 401--17, 1996.

\bibitem{Takizawa2024}
R.~Takizawa, Y.~Ohmura, and Y.~Kuniyoshi, ``Gaze-guided task decomposition for imitation learning in robotic manipulation,'' in \emph{Proc. {IEEE}/{RSJ} Int. Conf. Intell. Robots Syst.}, 2025.

\bibitem{Zhang2018}
T.~Zhang, Z.~McCarthy, O.~Jow, D.~Lee, K.~Goldberg, and P.~Abbeel, ``Deep imitation learning for complex manipulation tasks from virtual reality teleoperation,'' in \emph{Proc. {IEEE} Int. Conf. Robot. Automat.}, 2018.

\bibitem{Oquab2023}
M.~Oquab, \emph{et~al.}, ``{DINO}v2: Learning robust visual features without supervision,'' \emph{Trans. Mach. Learn. Res.}, 2024.

\bibitem{Zhao2020}
H.~Zhao, L.~Jiang, J.~Jia, P.~H. Torr, and V.~Koltun, ``Point transformer,'' in \emph{Proc. {IEEE} Int. Conf. Comput. Vis.}, 2021.

\bibitem{Dosovitskiy2020}
A.~Dosovitskiy, \emph{et~al.}, ``An image is worth 16x16 words: Transformers for image recognition at scale,'' in \emph{Proc. Int. Conf. Learn. Represent.}, 2021.

\bibitem{He2015}
K.~He, X.~Zhang, S.~Ren, and J.~Sun, ``Deep residual learning for image recognition,'' in \emph{Proc. {IEEE} Conf. Comput. Vis. Pattern. Recognit.}, 2015.

\end{thebibliography}
\end{document}